\documentclass[conference]{IEEEtran}
\IEEEoverridecommandlockouts
\usepackage{cite}
\usepackage{amsmath,amssymb,amsfonts}
\usepackage{algorithmic}
\usepackage{graphicx}
\usepackage{textcomp}
\usepackage{xcolor}
\usepackage{pdfpages}
\usepackage{tikz}
\def\BibTeX{{\rm B\kern-.05em{\sc i\kern-.025em b}\kern-.08em
    T\kern-.1667em\lower.7ex\hbox{E}\kern-.125emX}}
\newcommand\copyrighttext{%
    \footnotesize This work has been submitted to the IEEE for possible publication. Copyright may be transferred without notice, after which this version may no longer be accessible.}

\newcommand\copyrightnotice{%
\begin{tikzpicture}[remember picture,overlay]
\node[anchor=south,yshift=10pt] at (current page.south) {\fbox{\parbox{\dimexpr0.75\textwidth-\fboxsep-\fboxrule\relax}{\copyrighttext}}};
\end{tikzpicture}%
}
\begin{document}

\title{OpenConvoy: Universal Platform for Real-World Testing of Cooperative Driving Systems\\
\thanks{This work supported in part by the National Science Foundation under Grant CNS-1932037.
Article processing charges were provided in part by the UCF College of Graduate Studies Open Access Publishing Fund.}
}

\author{\IEEEauthorblockN{Burns, Owen}
\IEEEauthorblockA{\textit{College of Engineering and Computer Science} \\
\textit{University of Central Florida}\\
Orlando, United States \\
https://orcid.org/0009-0007-1826-373X} \\
\IEEEauthorblockN{Yaser Fallah}
\IEEEauthorblockA{\textit{College of Electrical and Computer Engineering} \\
\textit{University of Central Florida}\\
Orlando, United States \\
Yaser.Fallah@ucf.edu}
\and
\IEEEauthorblockN{Hossein Maghsoumi}
\IEEEauthorblockA{\textit{College of Electrical and Computer Engineering} \\
\textit{University of Central Florida}\\
Orlando, United States \\
hossein.maghsoumi@ucf.edu} \\
\IEEEauthorblockN{Israel Charles}
\IEEEauthorblockA{\textit{College of Engineering and Computer Science} \\
\textit{University of Central Florida}\\
Orlando, United States \\
israel.charles@ucf.edu} \\
}

\maketitle

\begin{abstract}
Cooperative driving, enabled by communication between automated vehicle systems, promises significant benefits to fuel efficiency, road capacity, and safety over single-vehicle driver assistance systems such as adaptive cruise control (ACC). However, the responsible development and implementation of these algorithms poses substantial challenges due to the need for extensive real-world testing. We address this issue and introduce OpenConvoy, an open and extensible framework designed for the implementation and assessment of cooperative driving policies on physical connected and autonomous vehicles (CAVs). We demonstrate the capabilities of OpenConvoy through a series of experiments on a convoy of multi-scale vehicles controlled by Platooning to show the stability of our system across vehicle configurations and its ability to effectively measure convoy cohesion across driving scenarios including varying degrees of communication loss.
\end{abstract}

\copyrightnotice

\section{Introduction}
Rapid advancements in autonomous vehicle technology and vehicle to vehicle communication (V2V) have opened up new avenues for enhancing transportation efficiency, safety, and sustainability \cite{v2xadvances}. One promising application within this domain is cooperative driving, where cars receive movement information from all the vehicles traveling in front of it and can thus slow down or speed up in unison. This coordination enables shorter safe inter-vehicle distances, reducing emissions by reducing drag on the cars in the convoy and increasing highway throughput \cite{doi:10.1080/15472450.2020.1720673}.

Despite these potential benefits, developing safe cooperative driving policies remains a complex challenge. Because of the short inter-vehicle distances and human lives involved, thorough testing must be undertaken to assess the resilience of such algorithms to issues such as rough terrain, communication loss, and noisy sensor readings. While testing in simulation can be helpful, simulators remain limited in their ability to replicate these stochastic conditions and sim2real transfer is rarely a straightforward or cheap process as a result \cite{xu2023opencda}. This problem is only compounded by the vast diversity of vehicle configurations to which a single convoy algorithm may be applied in the course of real-world use \cite{75eb35f0-ef72-3076-a9bc-01464345aecc}. Thus, there is an urgent need for a unified platform for assessing the performance of arbitrary cooperative driving implementations on arbitrary autonomous vehicles in a consistent manner. In this paper, we address this problem and introduce OpenConvoy to fill that gap. The contributions of this paper are as follows:
\begin{itemize}
    \item We introduce OpenConvoy, an open and modular platform that streamlines the implementation and testing of arbitrary cooperative driving algorithms on arbitrary autonomous vehicles.
    \item Our method directly supports ROS1 and ROS2 and uses the MAVLink protocol, enabling its direct application to a wide variety of vehicle types and control stacks with minimal modification.
    \item We demonstrate the stability of our system by successfully employing it in testing Platooning with time-triggered communication and an all-predecessor following IFT on a convoy of multi-scale vehicles.
\end{itemize}

\section{Background}
\subsection{ACC}
Adaptive Cruise Control (ACC) is a pivotal technology in the evolution of automated driving systems. ACC uses sensors such as radar, LIDAR, and cameras to monitor the distance and relative speed of the vehicle ahead \cite{Yeong2021}. By automatically adjusting the vehicle's speed, ACC maintains a safe following distance, reducing the need for driver intervention in maintaining speed and distance. This technology enhances driver comfort and safety by responding to changes in traffic conditions more smoothly than traditional cruise control systems, but lacks string stability when employed by large numbers of consecutive vehicles, amplifying traffic waves \cite{gunter2019commercially}.

\subsection{Cooperative Driving}
Cooperative driving systems use V2V and V2X communication to enhance the situational awareness of cars in close proximity to each other sharing of information such as speed, acceleration, and braking in real time \cite{ADNANYUSUF2024100980}. These systems have been demonstrated to provide a number of benefits, in particular improved highway throughput \cite{doi:10.1177/154193121005402403}\cite{4019451}\cite{doi:10.3141/2324-08}\cite{doi:10.1080/15472450.2017.1404680} and fuel efficiency \cite{doi:10.1080/15472450.2019.1608441}\cite{doi:10.1080/15472450.2020.1720673}\cite{doi:10.1080/15472450.2020.1720673}\cite{electronics10192373}. This is achieved through enabling vehicles to safely travel in closer proximity to each other than they normally would without this heightened awareness \cite{Malik2021}, and implementations of this concept can typically be deconstructed along the lines of spacing policy, communication policy, and controller.

\begin{figure}[!t]
  \centering  \includegraphics[width=1\linewidth]{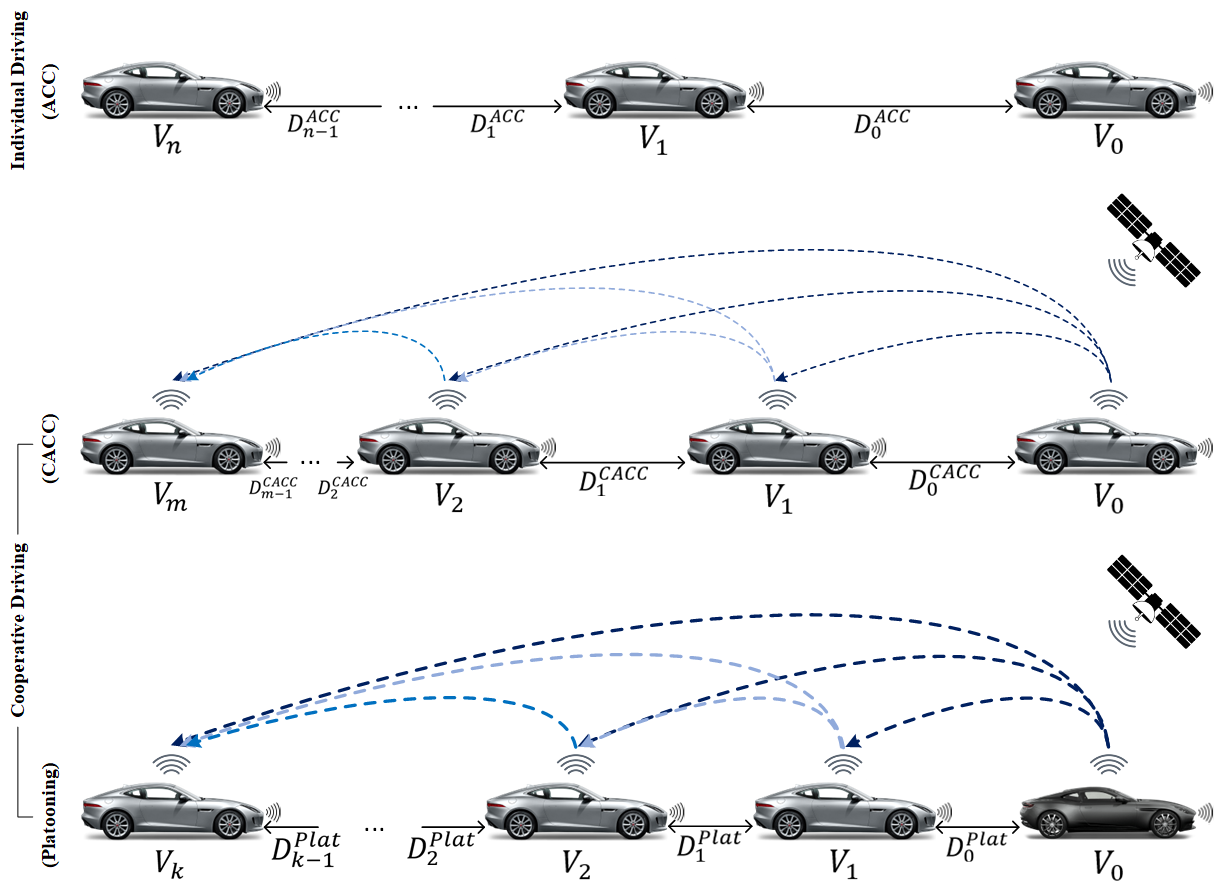}
  \vspace{-.5cm}
  \caption{ACC compared with CACC and Platooning, each using an all-predecessor IFT}
  \label{fig:ThreeinOne}
\end{figure}

\subsubsection{Spacing Policy}
In cooperative driving, the spacing policy is the optimization objective defining the physical structure of the platoon. Spacing policies can typically be categorized as either platooning or CACC, with the former using fixed following distances and the latter using fixed following times \cite{Nowakowski2015}.

\subsubsection{Communication Policy}
The communication policy defines the structure of V2V communications in a cooperative driving system, and includes the policy for sending messages (either time- or event-triggered \cite{10076814}) and the network topology that determines who receives them. Network topologies can be greatly varied, but are commonly either bidirectional or predecessor following with varying degrees of lookahead \cite{7055887}\cite{9837992}, visualized in comparison to ACC in Fig.~\ref{fig:ThreeinOne}.

\subsubsection{Controller}
Based on the information received from the other vehicles in the convoy and the objective of the spacing policy, the controller determines the motion to be executed at the vehicle level; these controllers can vary widely from simple linear controllers to advanced MPC implementations \cite{9625177}.

\begin{table*}[h]
    \caption{Cooperative Driving Implementations}
    \centering
    \begin{tabular}{ |p{1.5cm}||p{2cm}|p{2.5cm}|p{2.75cm}|p{3cm}|  }
     \hline
     \multicolumn{5}{|c|}{Cooperative Driving Implementations} \\
     \hline
     Reference       &Spacing Policy&IFT&Communication Trigger&Model-based Controller\\
     \hline
     \cite{7445860}&M-CACC&All BD&Time&Yes \\
     \hline
     \cite{10076814}&CACC&All PD&Event&Yes \\
     \hline
     \cite{9837992}&CACC&N-lookahead PD&Time&Yes \\
     \hline
     \cite{WIJNBERGEN2021104954}&M-CACC&1-lookahead PD&Time&No \\
     \hline
     \cite{9625177}&Platooning&All PD&Time&Yes \\
     \hline
     \cite{9447985}&Platooning&All PD&Event&Yes \\
     \hline
     \cite{WEI201796}&Platooning&Leader Following&Event&Yes \\
     \hline
     \cite{8574938}&CACC&1-lookahead PD&Event&Yes \\
     \hline
     \multicolumn{4}{l}{BD: Bidirectional, PD: Predecessor, M-CACC: modified CACC}
    \end{tabular}
    \label{fig:Implementations}
\end{table*}

\begin{figure}
    \centering
    \scalebox{0.5}{%
        \includegraphics[trim=8cm 0cm 0cm 0cm, clip]{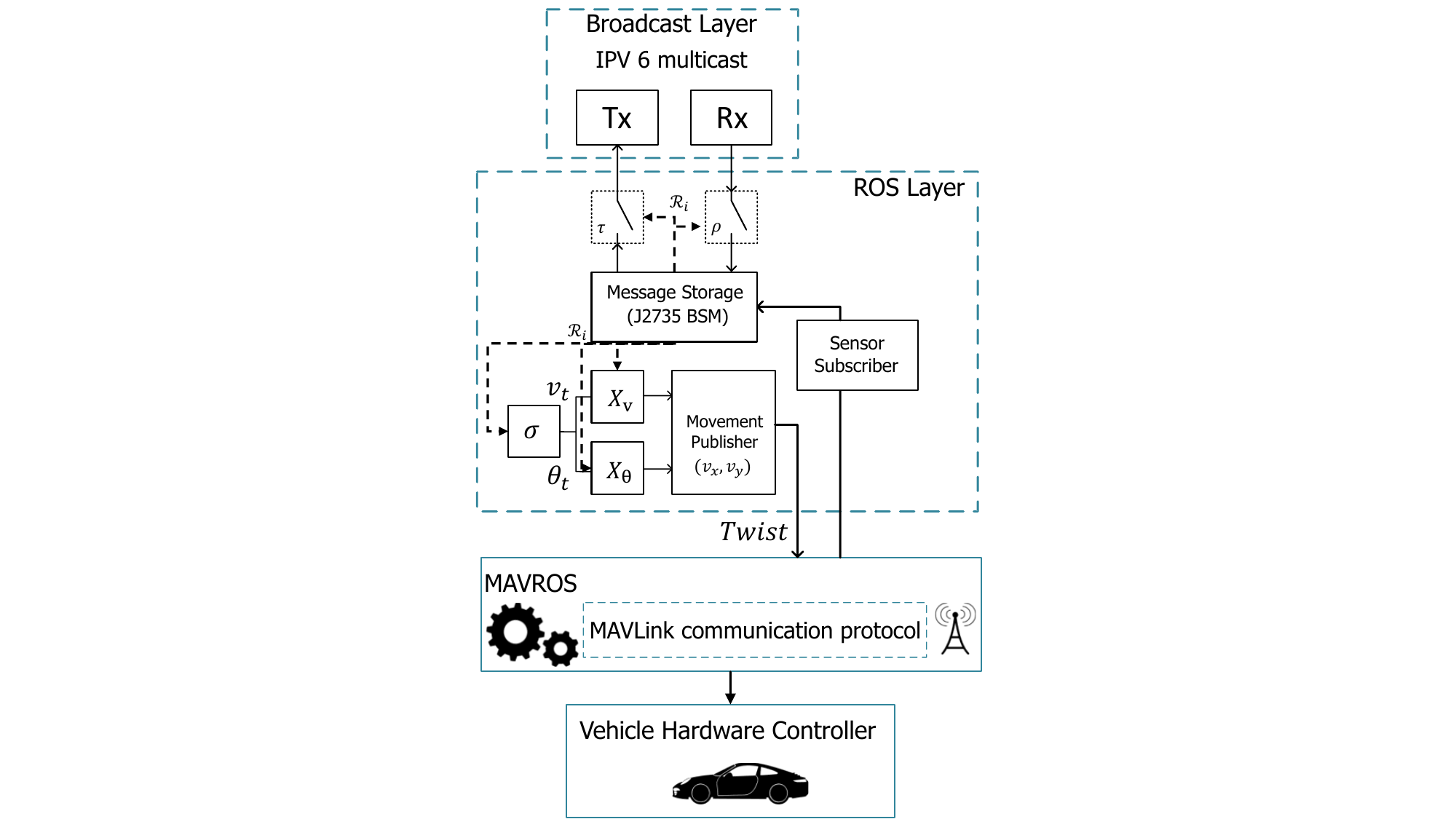}
    }
    \caption{Block diagram of OpenConvoy platform}
    \label{fig:block-diagram}
\end{figure}

Fig.~\ref{fig:Implementations} shows a number of different cooperative driving implementations broken down by their spacing policy, communication policy and controller. The vast majority of implementations use some form of MPC due to its flexibility, though \cite{WIJNBERGEN2021104954} uses a non-model based geometric controller to compute a desired following distance that varies quadratically with the speed of the vehicle. By contrast, \cite{7445860} uses model based control to alter vehicle speed to meet the desired following time and vehicle-specific speed profile, which is computed with respect to the motion of the entire platoon and the grade of the road to maximize fuel efficiency. Despite these stark differences, the remaining logic (how messages are broadcast, how the code communicates with the vehicle hardware, etc.) is functionally identical.

\section{Related Work} 
The cost and time associated with conducting field tests of autonomous driving has created the need for easy-to-use frameworks to simulate the various aspects of autonomous driving. Common robotics simulators such as Gazebo \cite{1389727} and Simulink \cite{MATLAB} are insufficient to meet this demand due to their granularity, making it difficult to specify environments large enough for effective autonomous vehicle testing. By contrast, simulators such as Carla \cite{dosovitskiy2017carla} and LGSVL \cite{rong2020lgsvl} have emerged which optimize tradeoff between the performance required for testing complex functionality and the detail required for the simulation to be an accurate indication of performance. Other simulators have emerged for more niche simulation needs, including CarSim \cite{10.5555/1628275.1628283}, which specializes in simulating accurate vehicle dynamics. Separately, traffic flow simulators such as SUMO \cite{SUMO2018} and Veins \cite{Sommer2019} to conduct high-level multi-vehicle simulations for scenarios which don't require vehicle-level detail.

However, these developments in autonomous driving simulation did not extend to cooperative driving, which requires both multi-vehicle co-simulation and vehicle-level dynamics, as well as the entirely novel ability to simulate vehicle to everything (V2X) or V2V communication. This spurred the development of dedicated co-simulation frameworks such as OpenCDA \cite{xu2021opencdaan} and AutoCastSim \cite{Cui_2022}, with V2Xverse \cite{liu2024collaborative} being developed later in part to expand the range of available environments and benchmarks. Outside of these standard frameworks, a large number have been developed for niche cooperative driving tasks, commercial use, or rapid prototyping\cite{Franke2015ACD}\cite{8891221}\cite{Bischoff2020WhatCC}\cite{duadas2018}\cite{9733618}.

Despite these advancements in simulation techniques, simulators are still insufficient to capture the full breadth of possibilities that may occur in the real world, and as such real-world testing is still a crucial requirement for the safe development and deployment of autonomous driving systems \cite{75eb35f0-ef72-3076-a9bc-01464345aecc}. To address this need, many of the noncooperative simulators (e.g. CARLA) mentioned previously include some functionality for sim2real transfer, though these methods are focused on machine learning-based systems and include transfer learning, knowledge learning, and robust reinforcement learning \cite{10242366}. Typical approaches to bridging the sim2real gap for cooperative driving primarily focus on creating open datasets for real-world V2V communication and creating more in-depth real-world benchmarks \cite{Xu_2023_CVPR}\cite{xu2023opencda} or creating more realistic simulations of sensor errors \cite{8628408}. Some work has been conducted in creating a platform for testing cooperative driving algorithms on physical hardware, but is restricted to a small set of hardware configurations and can't be scaled to larger vehicles for deployment-level testing \cite{8424491}.

\section{System Architecture}
\subsection{Problem specification}
In general, cooperative driving implementations differ in their choice of the following components:
\begin{enumerate}
    \item Communication Policy
    \item Spacing Policy
    \item Controller
\end{enumerate}
In order to facilitate arbitrary implementations of those components, we define cooperative driving implementations in general as follows, and the remainder of this section provides an overview of how the rest of the logic was implemented to facilitate compatibility across hardware and software stacks.

Let $i\in{0,1,...n}$ denote the vehicles of a convoy where vehicle $0$ is the leader. For each vehicle $i$, let $\mathcal{R}^i$ be a data structure containing the most recent states of the vehicles in the convoy it receives messages from according to the IFT, including itself and its own current state. Concretely,
\begin{equation}
    \begin{aligned}
{\mathcal{R}_j}^i=\{{s_j}^{n-k}, {s_j}^{n-k+1},...,{s_j}^n\}
    \end{aligned}
\end{equation}
When vehicle $i$ receives a message from vehicle $j$, the IFT of the system determines whether the message from $j$ should be saved; this Rx gate is formulated as:
\begin{equation}
    \begin{aligned}
\rho:(i, j)\rightarrow\{0,1\}
    \end{aligned}
\end{equation}
Similarly, every $\mathcal{B}$ seconds vehicle $i$ determines whether it should broadcast its current state based on $\mathcal{R}^i$; this Tx gate is formulated as:
\begin{equation}
    \begin{aligned}
\tau:(\mathcal{R}^i)\rightarrow\{0,1\}
    \end{aligned}
\end{equation}
At each time step $t$, vehicle $i$ calculates its target speed $v_t$ and heading $\theta_t$ based on the spacing policy $\sigma$ as follows:
\begin{equation}
    \begin{aligned}
v_t, \theta_t = {\arg\min}_{v, \theta}[\sigma(v, \theta, \mathcal{R}^i)]
    \end{aligned}
\end{equation}
Finally, the controllers $\chi_v$ and $\chi_\theta$ each transform the target speed and heading into the applied speed and heading, which is then actuated on the vehicle.
\begin{equation}
    \begin{aligned}
v_a=\chi_v(v_t, \mathcal{R}^i) \\
v_t=\chi_\theta(\theta_t, \mathcal{R}^i)
    \end{aligned}
\end{equation}
This definition allows arbitrary definition of $\tau$, $\rho$, $\sigma$, $\chi_v$, and $\chi_\theta$ to cover the breadth of cooperative driving implementations without any ROS- or hardware-specific logic. The remaining components are explained below, with Fig.~\ref{fig:block-diagram} being a block diagram of the overall system.

\begin{figure*}[!t]
  \centering  \includegraphics[width=1\linewidth]{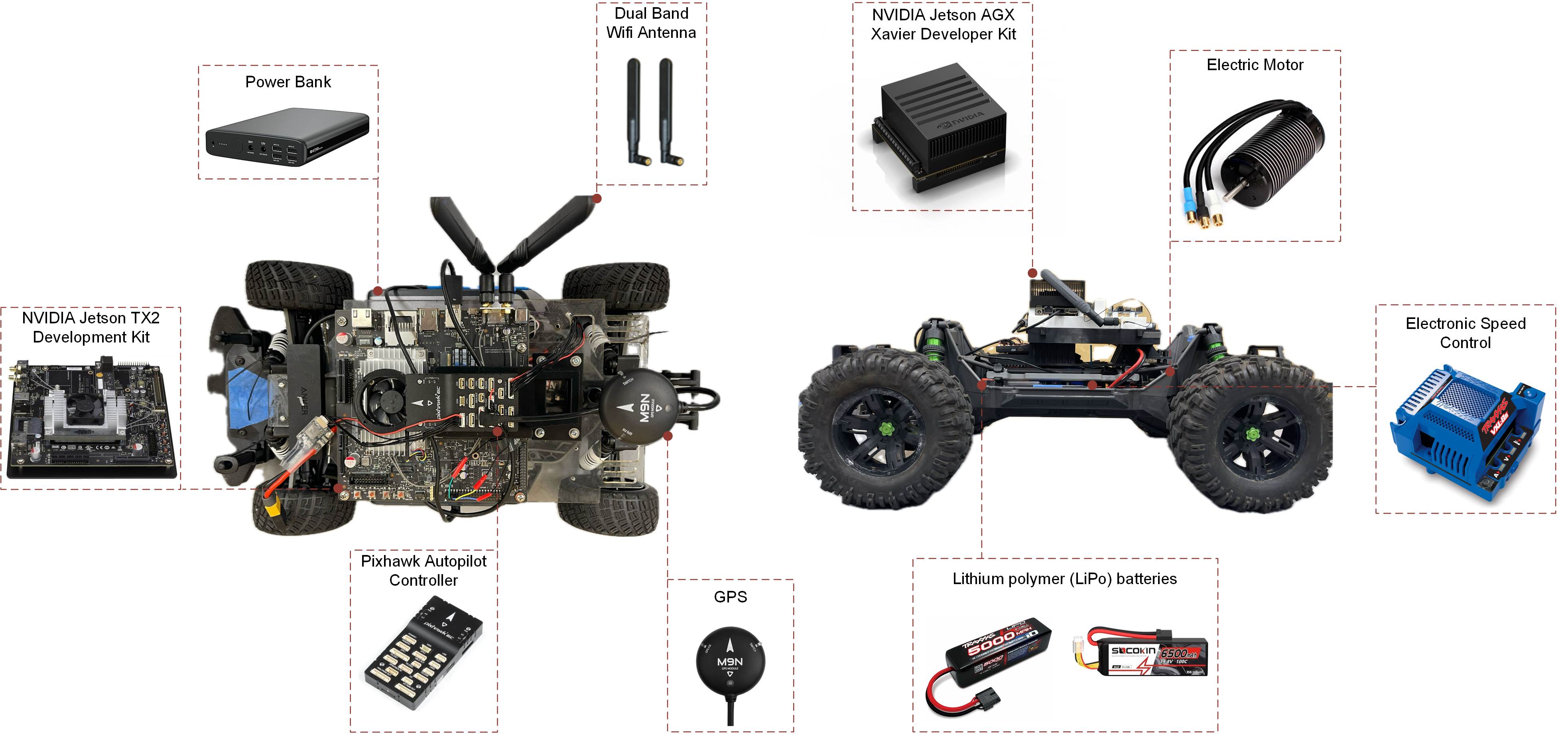}
  \vspace{-.5cm}
  \caption{Hardware Components for 1/10\textsuperscript{th} (left) and 1/6\textsuperscript{th} (right) Vehicles}
  \label{Vehicles_components}
\end{figure*}

\subsection{Hardware Interface}
To support the widest breadth of vehicle configurations possible, OpenConvoy uses MavRos \cite{mavros} to communicate with a flight controller running either PX4 \cite{7140074} or Ardupilot \cite{ardupilot}, which in turn communicates with the vehicle hardware. Because either firmware can run on a multitude of different flight controllers, and most flight controllers, ESCs, and GPS receivers are interchangable, OpenConvoy maintains near-universal compatibility with lab-scale autonomous vehicles.

\subsection{V2V Communication}
At the core of OpenConvoy’s communication architecture is the use of the SAE J2735 Basic Safety Messages (BSM) \cite{ahmed2019bsm} standard. This choice ensures that our platform can seamlessly interact with other cooperative driving systems adhering to the same standard, and that the vehicle states stored in $\mathcal{R}^i$ are in a widely accepted format, facilitating interoperability and broadening the scope of applications. BSMs encapsulate essential vehicle state information such as position, speed, heading, and acceleration, enabling real-time data exchange crucial for cooperative maneuvers.

To transmit these BSMs, OpenConvoy employs IPv6 multicast communication. This approach allows for efficient one-to-many communication, ensuring that messages are delivered to all vehicles within a defined group simultaneously while avoiding network configuration pitfalls that may block IPv4 broadcasts. This reduces network congestion and latency, critical factors in the dynamic environment of cooperative driving. Moreover, IPv6’s large address space supports the scalability required for large-scale deployments, accommodating the growing number of connected vehicles.

\begin{figure*}[!t]
  \centering  \includegraphics[width=1\linewidth]{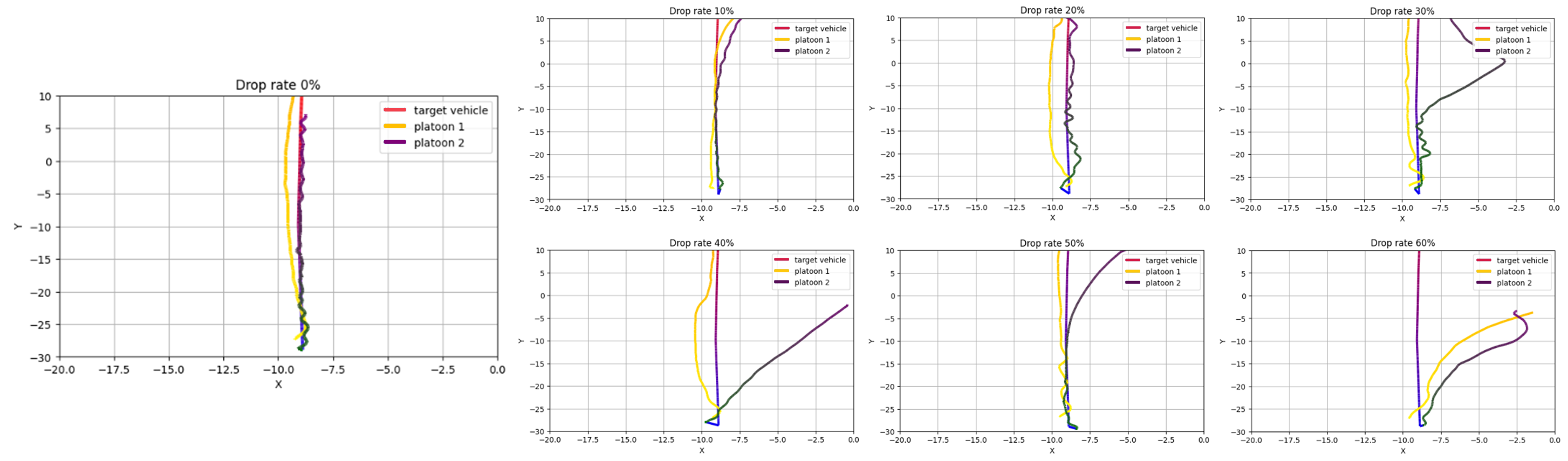}
  \vspace{-.5cm}
  \caption{Vehicle Trajectories}
  \label{Vehicle_Trajectories}
\end{figure*}

\begin{figure}[!ht]
  \centering  \includegraphics[width=0.93\linewidth]{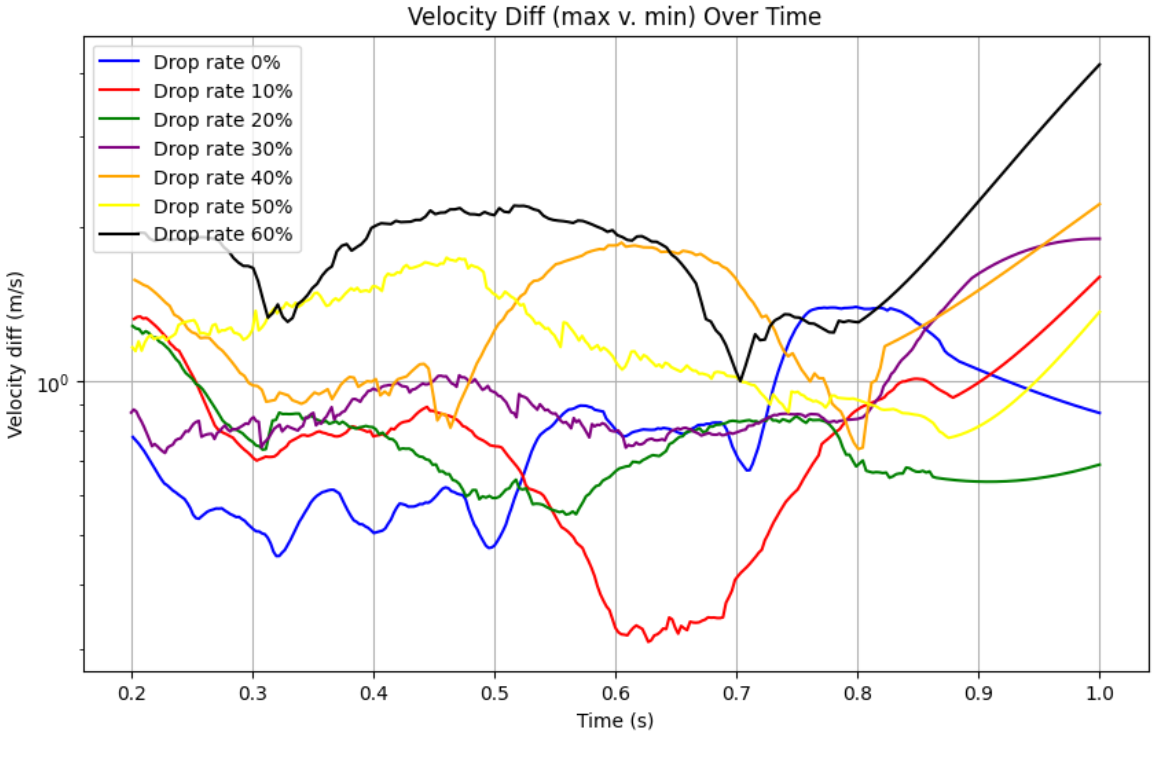}
  \vspace{-.5cm}
  \caption{Max inter-vehicle velocity difference over time}
  \label{VelDiff_Time}
\end{figure}

\begin{figure}[!ht]
  \centering  \includegraphics[width=0.93\linewidth]{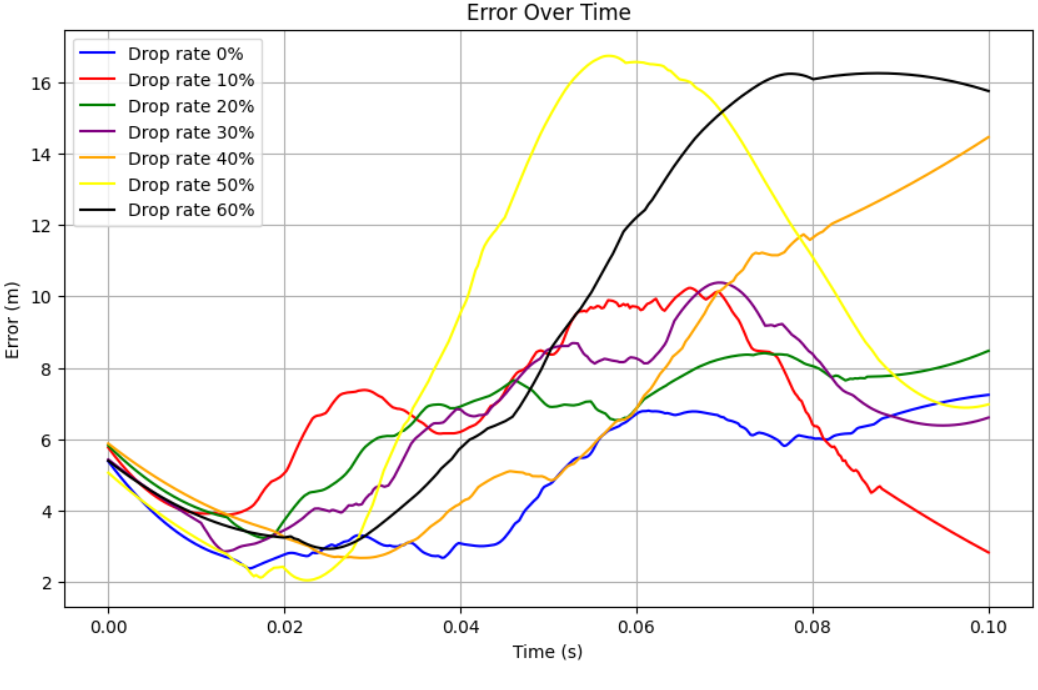}
  \vspace{-.5cm}
  \caption{Max inter-vehicle platooning error over time}
  \label{Error_Time}
\end{figure}

\begin{figure}[!ht]
  \centering  \includegraphics[width=0.93\linewidth]{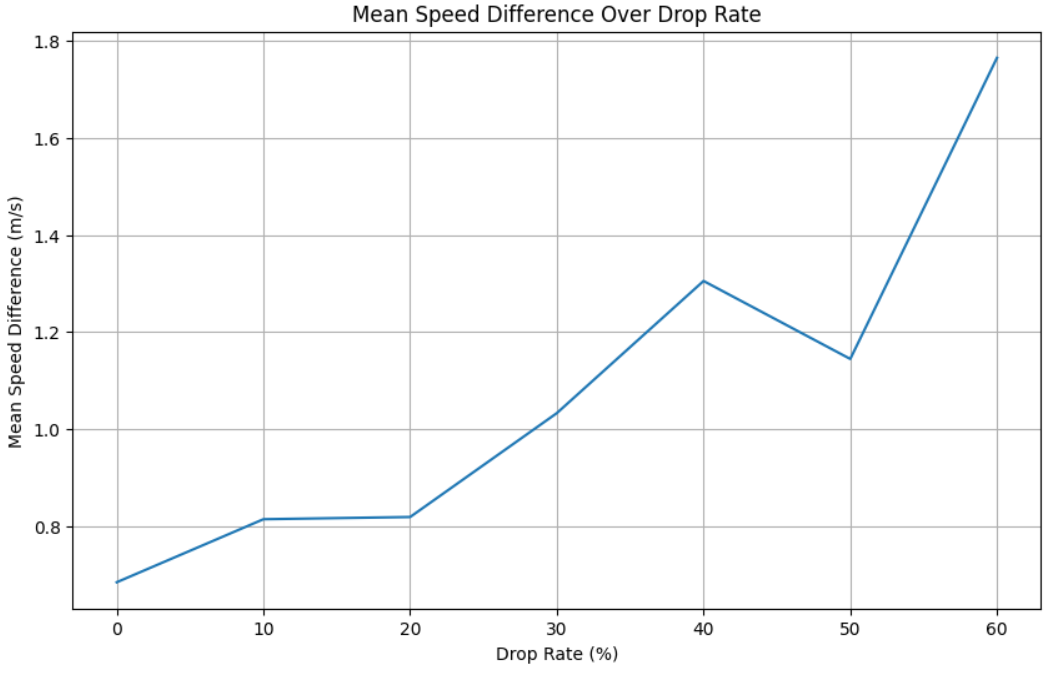}
  \vspace{-.5cm}
  \caption{Mean velocity error over drop rates}
  \label{MeanSpeedDiff_DropRate}
\end{figure}

\begin{figure}[!ht]
  \centering  \includegraphics[width=0.93\linewidth]{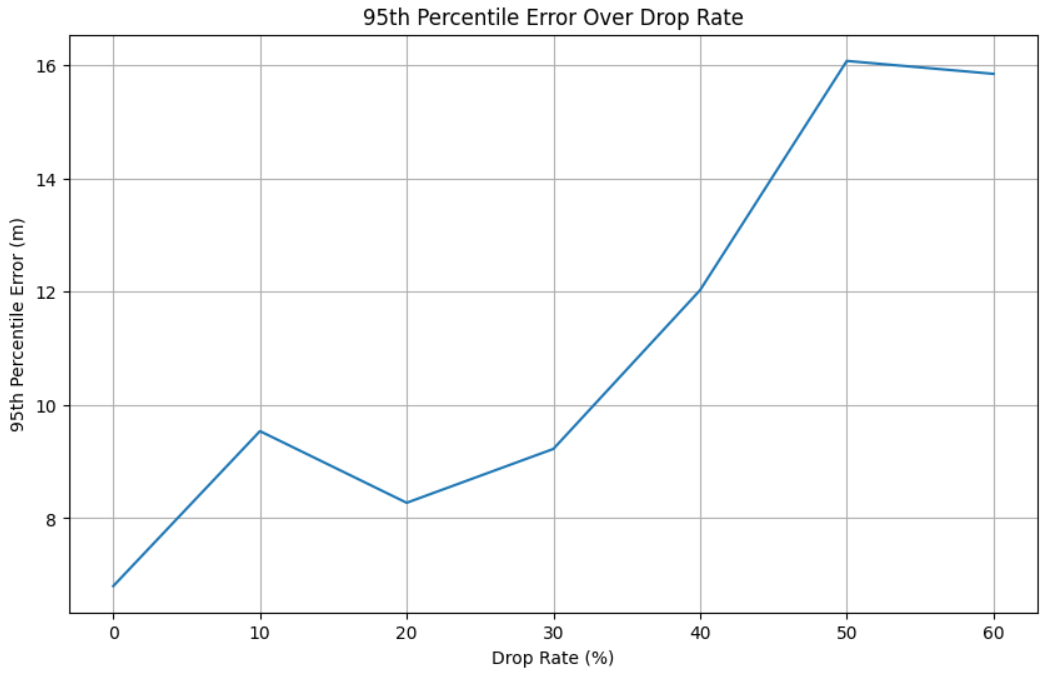}
  \vspace{-.5cm}
  \caption{95\textsuperscript{th} percentile platooning error over drop rates}
  \label{95PerError_Drop}
\end{figure}

\section{Experiments}
We conducted multiple experiments on a string of three small-scale vehicles, using OpenConvoy to implement platooning with an all-predecessor following IFT and time-triggered communication with a non-model based controller. We tested the performance of the convoy with varying degrees of communication loss in order to simulate the environment of a crowded network or imperfect transmission equipment.

\subsection{Hardware Setup}
For our experiments, we assembled three small-scale vehicles: two at a 1/10\textsuperscript{th} scale and one at a 1/6\textsuperscript{th} scale, each with different motors, servos, and speed controllers. Fig.~\ref{Vehicles_components} illustrates the hardware components of two of these vehicles, highlighting their design and setup. For each of the vehicles, we used a PixHawk 6C flight controller and M9N GPS receiver, with the PixHawk directly connecting to the ESC and steering servo in each vehicle. We used PX4 in all cases, though ArduPilot could be used interchangably due to its support for Mavros. One of the 1/10\textsuperscript{th} scale vehicles used ROS 1 and Python 2 on an Nvidia Jetson TX2 while both other vehicles used an Nvidia Jetson AGX Xavier with ROS 2 and Python 3.

\subsection{Implementation Details}
Using all-predecessor following, we let $\tau$ always return $1$ and $\rho$ return $1$ if the broadcaster is a predecessor in the platoon. We define $\sigma$ to be the sum of costs of the distance (in the plane arrived at by performing a local ENU transformation on all relevent GPS coordinates) between the point representing the correct following distance and the projected point one time step forwards for the ego vehicle for each predecessor of the ego vehicle in the platoon. For $\chi_v$ and $\chi_\theta$ we use a PD \cite{1453566} and Stanley controller \cite{4282788} respectively.
 
The experiments were conducted with these three vehicles under a controlled setting. The leader vehicle's trajectory was designed to mimic a realistic highway driving scenario, with a time-varying speed starting at 1 m/s and jumping to 2 m/s before returning to 1 m/s. This setup ensured that the following vehicles had to adapt to sudden speed changes, providing valuable data on the efficacy of our control algorithms in maintaining desired inter-vehicle distances and overall platooning stability. Higher speeds were ruled out due to the low top speed of our lowest powered following vehicle, which struggled to cope with the additional weight of the power bank and companion computer.

\subsection{Evaluation Metrics}
In evaluating the performance of our platooning experiments, we utilized two key metrics: Platooning Error \cite{shladover2015cooperative} and Speed Difference \cite{razzaghpour2021impact}. These metrics provided a comprehensive assessment of our control algorithms' effectiveness in maintaining desired distances and ensuring traffic flow stability.

\subsubsection{Platooning Error}
As introduced in \cite{shladover2015cooperative}, Platooning Error is defined as the absolute value of the difference between the actual distance gap and the desired distance gap (15 meters) in meters. To capture a statistical sense of worst-case behavior and account for the characteristics of wireless networks, we chose the 95\textsuperscript{th} percentile of the error’s absolute value. This metric allows us to evaluate how well the vehicles maintain the desired inter-vehicle spacing, even under varying conditions.

\subsubsection{Speed Difference}The variation between the highest and lowest speeds among all vehicles in the string at each time step serves as an effective indicator of traffic flow. As outlined in \cite{razzaghpour2021impact}, this metric, referred to as Speed Difference, is utilized to assess the performance of the platooning system. By evaluating the speed difference, we can determine how consistently the vehicles in the string are moving in relation to each other, which is crucial for maintaining smooth and efficient traffic flow.

These metrics together provide a detailed evaluation of the platooning system's performance, highlighting both the accuracy in maintaining desired distances and the stability of vehicle speeds within the platoon.

\subsection{Results}
Fig.~\ref{Vehicle_Trajectories} shows the vehicle trajectory for the lead vehicle and follower vehicles in scenarios with variable packet error rates (PER) ranging from 0\% to 60\%. As shown in this figure, with a drop rate between 0\% and 20\%, follower vehicles 1 and 2 follow the lead vehicle with good accuracy. In the drop rate range of 30\% to 50\%, although the first follower vehicle responds well and follows the lead vehicle with minor errors, the second follower vehicle experiences an increased error rate at times. At a drop rate of 60\%, the first follower vehicle loses the path of the lead vehicle, which leads to a path error for the second follower vehicle as well.

Fig.~\ref{VelDiff_Time} shows the difference between the maximum speed value and the minimum speed value at different time intervals for variable PER scenarios. It can be seen that this difference increases with the rise in drop rate in most time intervals.

Fig.~\ref{Error_Time} illustrates the platooning error over time intervals for various PER scenarios. Although, in most time intervals, the error for drop rate values between 0\% and 30\% is generally compact and close to each other, the error rate in some time intervals for drop rate values of 40\%, 50\%, and 60\% is significantly higher than the error values for drop rates between 0\% and 30\%.

For a better comparison of the maximum and minimum speed differences in platooning, the average speed difference over variable PER is shown in Fig.~\ref{MeanSpeedDiff_DropRate}\footnote{The noise in this and Fig.~\ref{95PerError_Drop} is due to our small sample size; we are still in the process of conducting additional trials and will update the figures accordingly.}. This figure clearly shows an increasing trend in the difference as the drop rate increases.

Additionally, the 95\textsuperscript{th} percentile error over different values of PER is shown in Fig.~\ref{95PerError_Drop}, highlighting the impact of increased PER on platooning performance. These results are directly consistent with the experiments carried out in simulator in \cite{9625177}, demonstrating the ability of OpenConvoy to provide an effective platform to bridge the sim2real gap for cooperative driving.

\section{Conclusion}
We present OpenConvoy to address the lack of a platform for easily implementing cooperative driving algorithms on platoons of real vehicles, reducing the work required to perform the kind of rigorous real-world testing which is required for the safe development and deployment of cooperative driving systems. We demonstrate the compatibility of OpenConvoy across multi-scale autonomous vehicles and its ability to rigorously assess the performance of cooperative driving implementations across varying communication landscapes, successfully replicating the results of previous simulator-based works. Future lines of work will include expanding testing to larger numbers of vehicles and implementing the methods of more papers, in particular creating an easily extensible basis for implementing MPC.

\bibliographystyle{IEEEbib}
\bibliography{strings,citations}

\end{document}